\documentclass[sigconf]{acmart}

\AtBeginDocument{%
  }

\setcopyright{acmcopyright}
\copyrightyear{2018}
\acmYear{2018}
\acmDOI{XXXXXXX.XXXXXXX}

\acmConference[Conference acronym 'XX]{Make sure to enter the correct
  conference title from your rights confirmation emai}{June 03--05,
  2018}{Woodstock, NY}
\acmPrice{15.00}
\acmISBN{978-1-4503-XXXX-X/18/06}

\usepackage{algorithmic}
\usepackage{graphicx}
\usepackage{textcomp}
\usepackage{xcolor}
\usepackage{booktabs,tabularx,ragged2e}
\usepackage{cleveref}
\usepackage{tikz}
\usepackage{pgfplots}
\usepackage{soul}
\usepackage{array,multirow,graphicx}
\usepackage{float}
\usepackage{flushend}
\usepackage{subfig}
\usepackage{tabu}

\usepackage{makecell}

\usepackage{times}
\usepackage{url}
\usepackage{graphicx}
\usepackage{algorithmic}
\begin{document}

\title{On the Consistency and Robustness of Saliency Explanations for Time Series Classification}
\author{Chiara Balestra}
\authornote{Both authors contributed equally to this research.}

\affiliation{%
	\institution{TU Dortmund}
	\country{Germany}}
\email{chiara.balestra@tu-dortmund.de}
\author{Bin Li}
\authornotemark[1]

\affiliation{%
	\institution{TU Dortmund}
	\country{Germany}}
\email{bin.li@tu-dortmund.de}

\author{Emmanuel Müller}
\affiliation{%
	\institution{TU Dortmund}
	\country{Germany}
}
\email{emmanuel.mueller@tu-dortmund.de}

\renewcommand{\shortauthors}{Balestra and Li et al.}

\begin{abstract}
	Interpretable machine learning and explainable artificial intelligence have become essential in many applications. The trade-off between interpretability and model performance is the traitor to developing intrinsic and model-agnostic interpretation methods. Although model explanation approaches have achieved significant success in vision and natural language domains, explaining time series remains challenging. The complex pattern in the feature domain, coupled with the additional temporal dimension, hinders efficient interpretation. Saliency maps have been applied to interpret time series windows as images. However, they are not naturally designed for sequential data, thus suffering various issues.\par 
	This paper extensively analyzes the consistency and robustness of saliency maps for time series features and temporal attribution. Specifically, we examine saliency explanations from both perturbation-based and gradient-based explanation models in a time series classification task. Our experimental results on five real-world datasets show that they all lack consistent and robust performances to some extent. By drawing attention to the flawed saliency explanation models, we motivate to develop consistent and robust explanations for time series classification. 
	
\end{abstract}


\maketitle

\section{Introduction}
The spread of machine learning techniques to safety-critical applications has raised major concerns about model interpretability. Although shallow models are inherently easily interpretable and broadly applied, they often lack accuracy compared to more sophisticated prediction models whose predictions' interpretation is far from obvious. This trade-off between intepretability and model performance has increased the interest in post-hoc explanations techniques, ideally model-agnostic, straightforward, and robust. Saliency explanation recently gained traction and succeeded in various computer vision~\cite{pillai2022consistent,shrikumar2017learning} and natural language processing tasks~\cite{tsai2022faith,russell2019efficient}; However, explaining time series models still faces challenges.     Saliency maps provide pixel importance scores that can be easily interpreted, visually identifying the most relevant areas for the given task; A notable example is image classification, where saliency maps can visually highlight influential areas to the predicted label, making them valuable methods for explaining predictions. The importance scores are assigned using various techniques and are mainly divided into gradient-based and perturbation-based methods. Additionally, Shapley values do not fall into these two categories but provide one approach to getting saliency maps. The structure of time series data has generated interest in directly applying saliency maps to obtain meaningful explanations for classification models~\cite{ismail2020benchmarking}. However, images and time series represent fundamentally different types of data. On the one hand, the temporal dependency in time series leads to time-dependent changes in feature attribution. On the other hand, explanation approaches are often not directly applicable to time series models with recurrent- or attention-based components~\cite{choi2016retain,sarthak}. Although several approaches try to treat data windows of time series as images and apply vision explanation methods~\cite{ismail2020benchmarking,pan2021two}, the quality of such explanations is often questionable. We address two main issues that arise when using saliency maps for explaining time series data predictions; We introduce two concerns, i.e., the \emph{consistency} and the \emph{robustness} of saliency maps on time series, and show in the experiments that the typical attribution approaches used for time series are neither robust nor consistent. 

\begin{figure}[!t]
	\centering\includegraphics[width=\linewidth]{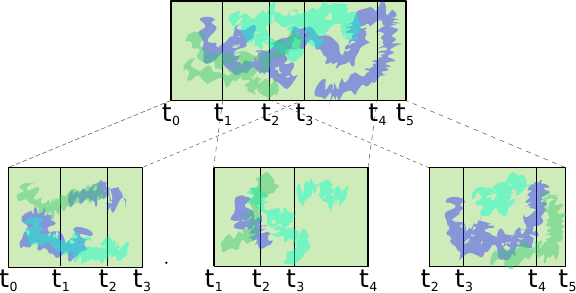}\caption{\label{fig:fig_inconsistency} Representation of inconsistencies among saliency maps of adjacent sliding windows ($x$-axis: time, $y$-axis: features, color: importance attribution). The saliency map in the top shows overall attribution from $t_0$ to $t_5$. The bottom row shows the saliency maps relative to three adjacent sliding windows (from left to right, in the interval $[t_0,t_3]$, $[t_1,t_4]$ and $[t_2,t_5]$). It is easy to spot that the color distributions in the overlapping window $[t_2,t_3]$ in the four cases are different.}    
\end{figure}

\begin{figure*}[!t]
	\centering
	\includegraphics[width=\textwidth]{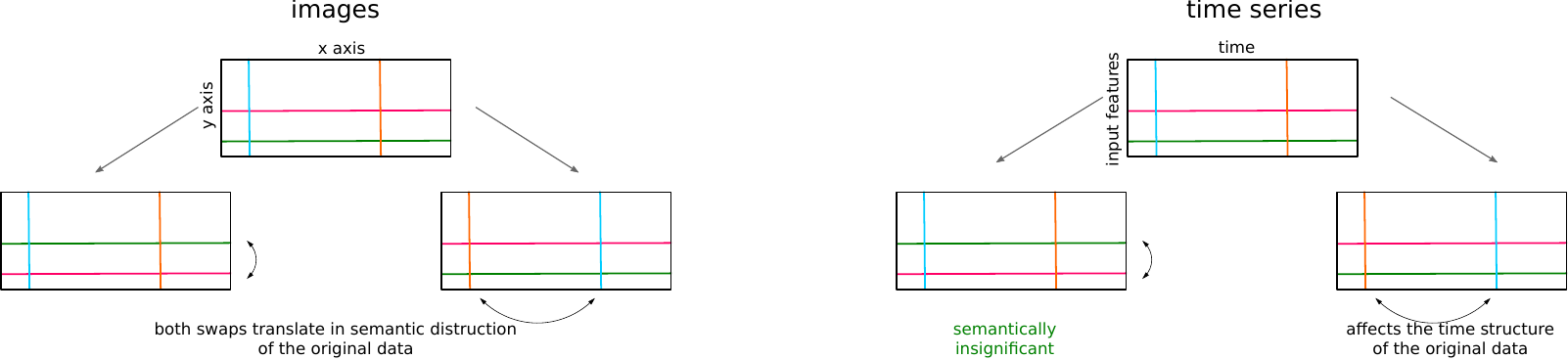}
	\caption{Representative drawings of time series data as $2D$ data frames. The position of a pixel in an image is defined by row and column numbers, while in time series data, by input variables and timestamps. Swaps of rows and columns of pixels in images may affect the semantic meaning of the entire data frame; For time series, this may happen only when swapping observations at different timestamps.}
	\label{fig:swaps}
\end{figure*}

    Deep time series analysis models usually consider sliding windows as basic input units to capture temporal information. Analog to saliency maps for visualizing image pixel importance, similar saliency maps are generated for time series frames with \emph{feature-time pixel} importance. Current research on explanation approaches for time series data can be classified into two categories. The first category contains methods treating sliding windows as \emph{frames} of images and applying classical image explanation methods, e.g., SHAP~\cite{SHAP}, LIME~\cite{ribeiro2016should}, and DeepLIFT~\cite{shrikumar2017learning}. Those methods extract local feature information while disregarding the time structure typical of the time series data. The second category contains methods considering the time dimension as an additional feature for joint explanation~\cite{bento2021timeshap,ismail2020benchmarking}. Overall, explanation methods on time series data consider the time dimension either jointly with the other features~\cite{pan2021two} or separately~\cite{ismail2020benchmarking}, through sequentially considering time and features. In both cases, the explanation is limited to one single \emph{frame} (i.e., sliding window). Hence, both categories reveal insufficient for interpreting the overall temporal information over a long time span. We claim that explanations over the intersection of sliding windows should exhibit \emph{consistent} behaviors to identify such a flaw in current time series saliency explanations. We admit that in adjacent sliding windows, different temporal contexts may lead to different absolute feature attribution. Therefore, we pursue \emph{consistency} relative attribution in local sub-windows. \Cref{fig:fig_inconsistency} illustrates the meaning of \emph{consistency} among overlapping time windows. \par
In addition to the saliency explanation consistency, the robustness of saliency maps against feature perturbation is another essential factor in ensuring explanation quality. In images, the semantic meaning of columns and rows is equivalent, while in time series, the time structure makes time series data semantically different and introduces dependence among the observations in the various timestamps; In images, swaps of both rows and columns of pixels affect the semantic structure of the original data, while in time series, only the swaps affecting the temporal orders of the observations are semantically meaningful. In contrast, the order in which input values are collected has no effects. The phenomenon is illustrated in Figure~\ref{fig:swaps} where the $x$-axis corresponds to the time, and the $y$-axis to the input variables.     When saliency maps are applied to time series, the salient features should be insensitive to the order of input features. When feature columns in the time series frame are swapped, important areas in the saliency map should stay salient in the corresponding swapped areas. We call this the \emph{robustness} of saliency explanation. \par
This paper studies the \emph{consistency} and \emph{robustness} of saliency explanation in the time series classification task. We examine saliency explanations from popularly used perturbation- and gradient-based approaches~\cite{suresh2017clinical,sundararajan2017axiomatic} on multiple deep classification models~\cite{hochreiter1997long,lea2017temporal,vaswani2017attention}. We show on five real-world datasets that the studied saliency explanation suffers from consistency and robustness issues. These preliminary results underline the encountered problems as a motivating example of further research on developing robust and consistent saliency explanations for time series.

\section{Related work}
Intrinsic and post-hoc explanations have transformed into a core topic for machine learning research; The interest in increasing the explainability of the methods embraces the entire Machine Learning and AI communities~\cite{shrikumar2017learning,strumbelj}. Gradient (GRAD) was introduced in Baehrens et al.~\cite{baehrens2010explain} in 2010 as a post-hoc model agnostic interpretation tool able to explain nonlinear classifiers at a local level; The provided explanations measure how each data point has to be moved to change the predicted label~\cite{baehrens2010explain}. The local scores derive from the direct computation of the local gradients (or their estimations) for the given model. Similarly, Integrated Gradients~\cite{sundararajan2017axiomatic} is also a gradient-based feature importance attribution method and builds up on ~\cite{baehrens2010explain} and two axioms, i.e., the \emph{sensitivity} and \emph{implementation invariance}. Finally, SHAP~\cite{SHAP} represents a successful attempt to introduce Shapley values in machine learning; Lundberg et al.~\cite{SHAP} use Shapley values to assign importance scores to features for local explanations of black-box models' predictions; The Shapley values' approximations are based on the computations of the gradient of the model predictions.\par
Although progress is not neglectable, the explanations provided by the most recent works are mostly not quantitatively evaluable, thus still raising trust issues in users~\cite{zhang2019should}. Few recent works focus on the quality of the explanation methods; Dombrowski et al.~\cite{dombrowski2019explanations} showed that explanations for image classification are non-robust against possible visually hardly detectable manipulations. \par
Explanation methods appear for most of the machine learning techniques with different strengths. Time series represents one niche data type where most implemented methods still lack explanations. One of the reasons for the poor literature on the explainability of time series data is the additional time-dependent structure. Explanations often reduce to applications of model-agnostic post-hoc explanations for general data samples to time-dependent data; The time structure is often disregarded, and the timestamps are treated as independent samples on which the model is learned~\cite{bento2021timeshap,schlegel2019towards}. Another thread of approaches using attention-based models obtains time-dependent explanations by attention weights~\cite{kaji2019attention,song2018attend,choi2016retain}. The acquired feature and time attribution to the prediction can be visualized in saliency maps, which are initially implemented for images~\cite{bach2015pixel}, and are a current trend in obtaining explanations for importance scores of timestamps and features; Among them, we find gradient-based~\cite{baehrens2010explain,sundararajan2017axiomatic,smilkov2017smoothgrad,shrikumar2017learning} and perturbation-based feature importance scores~\cite{zeiler2014visualizing,suresh2017clinical}. Ismail et al.~\cite{ismail2020benchmarking} pointed out how these methods often suffer from a lack of understanding of the time-features structures, either allowing to achieve only good performances at a time level or the features level; The authors propose an alternative two-step approach to saliency explanations for time series, where the time structure is first considered, and the importance of the features are considered only in the second step. The explanation quality of such a method is still under-studied in the time series domain. \par

\section{Open issues on saliency explanations}
This section formally defines the consistency and robustness of the saliency explanation for time series classification. \par 
We indicate with $X =(X_1,\ldots, X_N)$ a multivariate $N$-dimensional discrete time series where $X_i$ is the $i$-th univariate dimension; $t_{0}$ is the first timestamp on which the time series is defined. For each timestamp $t_k > t_{0}$, $X(t_k)$ is a $N$-dimensional vector of real values, i.e., $X(t_k) \in \mathbb{R}^N$. We study the consistency and robustness of saliency explanations for classification models trained on time series data. We draw upon the concept of \emph{consistency} proposed by Pillai et al.~\cite{pillai2022consistent}, and define \emph{consistency} of saliency explanations over adjacent sliding windows of time series. Additionally, regarding \emph{robustness}, we consider the influence of swaps of \emph{features}, i.e., of input variables observations, in explanations using saliency maps.

\subsection{Consistency}
We define \emph{time windows} $\{w_s^d\}_{s\in \mathbb{N}}$ dependent on the window length $d\in\mathbb{N}$ and the starting timestamp $t_s$, i.e.,
\begin{equation}\label{eq:timewindows}
	w_s^d = \{t_{{s}}, \ldots, t_{s+d-1} \}.
\end{equation}
For each time window and given a fixed saliency map method assignation of importance score $S$, we get $S(w_s^d) = S_s^d$ a matrix in $\mathbb{R}^{N\times d}$ such that $(S_s^d)_{n,t}$ is the importance scores assigned to the input variable $X_n$ at time $t$. Saliency maps are transposed from image (pre)processing applications to explain time series classification predictions. 
We examine the consistency of saliency maps defined over overlapping windows. Given two windows $w_s^d$ and $w_{\bar s}^{\bar d}$ such that $|w_s^d \cap w_{\bar s}^{\bar d}| \neq \emptyset$ and the respective saliency maps $S_s^d$ and $s_{\bar s}^{\bar d}$, the saliency explanations are inconsistent at timestamp $t$, if $t, \bar t \in w_s^d \cap w_{\bar s}^{\bar d}$ such that 
\begin{equation}
	(S_s^d)_{n,t} > (S_{\bar s}^{\bar d})_{n,t} \text{ and } (S_s^d)_{n,\bar t} < (S_{\bar s}^{\bar d})_{n, \bar t},
\end{equation}
i.e., the importance scores assigned to features and timestamps are \emph{relatively} inconsistent among overlapping time windows. The phenomenon is illustrated in Figure~\ref{fig:fig_inconsistency}. The distribution of colors in the saliency maps represents the importance of the timestamps and input variables. The different cuts of the time windows (second row of plots in Figure~\ref{fig:fig_inconsistency}) are characterized by different color distributions than the original saliency map in the first row.  

\subsection{Robustness}
Although similarly structured, we mentioned that images and time series intrinsically include a different semantic meaning due to the time dependency. However, the time series explanation should be insensitive to the feature ordering. A saliency explanation is considered as \emph{robust} if the saliency changes accordingly when the features are swapped. We define the feature swapping operation on data window $w_s^d$ and observe the effect in the corresponding saliency explanation $S_s^d$. Concretely, we swap random pair of features $X_i$ and $X_j$ ($i\neq j$) in $w_s^d$ for all timestamps from $t_s$ to $t_{s+d-1}$. Their feature attribution in $S_s^d$ are $(S_s^d)_i$ and $(S_s^d)_i$. After features swapping, the data window is denoted by ${w^*}_s^d$, and the newly learned saliency explanation is ${S^*}_s^d$. $(S_s^d)_i$ corresponds to $({S^*}_s^d)_j$ while $(S_s^d)_j$ corresponds to $({S^*}_s^d)_i$. The saliency explanations are robust if $t_1, t_2 \in w_s^d \cap w_s^d$ such that
\begin{equation}
	(S_s^d)_{i,t_1} > (S_s^d)_{i,t_2} \text{ and } ({S^*}_s^d)_{j,t_1} > ({S^*}_s^d)_{j,t_2},
\end{equation}
i.e., important feature-time pixels maintain relative importance after swapping the feature of the data window. The phenomenon is illustrated in~\Cref{fig:swaps}.

\section{Experiments}\label{sec:experiments} 

\begin{figure*}[t]
	\centering
	\includegraphics[width=1.0\textwidth]{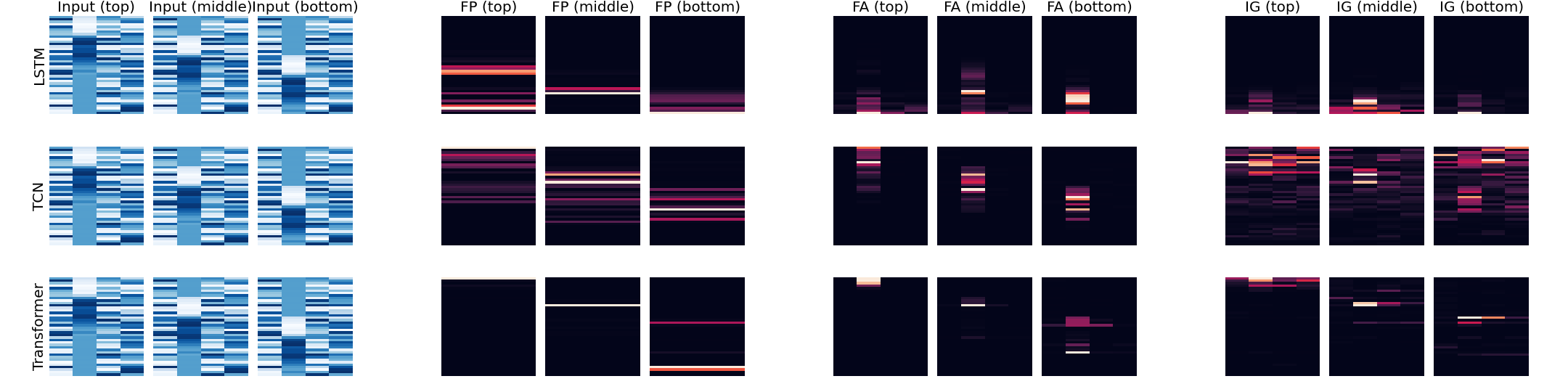}
	\caption{Saliency explanation of one data window from the IPD dataset: 
		The leftmost group in blue heatmaps denotes the variants of one input frame from the IPD dataset, where rows are timestamps and columns are features. The real data window is located in the frame's top, middle, and bottom third, and the rest elements are random noise. The saliency maps on the right side are acquired from three feature attribution algorithms: Feature Permutation (FP), Feature Ablation (FA), and Integrated Gradient (IG).}
	\label{fig:saliencymaps}
\end{figure*}

We perform experiments on time series classification on real-world datasets. We generate various types of explanations in the form of saliency maps for the predictions made by the model to examine their consistency and robustness. We incorporate artificial padding into the input sequences to precisely control the feature importance and simulate the sliding window mechanism commonly used in time series analysis tasks. This section presents our findings on identifying inconsistency and non-robust saliency explanations across multiple datasets.

\newcolumntype{b}{>{\arraybackslash}X}
\begin{table*}[!ht]
	\centering
	\caption{Consistency ranking analysis}
	\label{tab:kendalltau_std}
	
	\begin{tabularx}{\textwidth} {bb|bb|bb|bb}
		\toprule
		\multicolumn{2}{c}{}  
		& \multicolumn{2}{c}{Feature Permutation (FP)} & \multicolumn{2}{c}{Feature Ablation (FA)} & \multicolumn{2}{c}{Integrated Gradients (IG)} \\ 
		\cmidrule(lr){3-4}\cmidrule(lr){5-6}\cmidrule(lr){7-8}
		\multicolumn{2}{c}{}                   & $\tau$     & $\rho$    & $\tau$     & $\rho$    & $\tau$     & $\rho$      \\ \midrule
		\multirow{3}{*}{PD} & LSTM & $0.936\pm0.063$ & $0.967\pm0.035$ & $0.868\pm0.114$ & $0.921\pm0.092$ & $0.852\pm0.024$ & $0.968\pm0.012$\\& TCN & $0.475\pm0.143$ & $0.625\pm0.155$ & $0.247\pm0.200$ & $0.334\pm0.256$ & $0.060\pm0.148$ & $0.080\pm0.168$\\& Transformer & $0.040\pm0.136$ & $0.049\pm0.145$ & $-0.049\pm0.149$ & $-0.086\pm0.161$ & $0.026\pm0.138$ & $0.030\pm0.145$\\\midrule\multirow{3}{*}{WIN} & LSTM & $0.914\pm0.042$ & $0.961\pm0.023$ & $0.843\pm0.044$ & $0.908\pm0.030$ & $0.918\pm0.022$ & $0.971\pm0.009$\\& TCN & $0.445\pm0.164$ & $0.530\pm0.169$ & $0.220\pm0.159$ & $0.287\pm0.149$ & $0.052\pm0.200$ & $0.057\pm0.206$\\& Transformer & $0.150\pm0.192$ & $0.196\pm0.207$ & $-0.11\pm0.218$ & $-0.198\pm0.236$ & $0.107\pm0.180$ & $0.140\pm0.178$\\\midrule\multirow{3}{*}{IPD} & LSTM & $0.475\pm0.086$ & $0.633\pm0.101$ & $0.531\pm0.047$ & $0.725\pm0.045$ & $0.767\pm0.030$ & $0.921\pm0.019$\\& TCN & $0.001\pm0.083$ & $-0.002\pm0.112$ & $-0.004\pm0.075$ & $-0.031\pm0.095$ & $0.192\pm0.083$ & $0.277\pm0.114$\\& Transformer & $0.124\pm0.115$ & $0.167\pm0.157$ & $0.054\pm0.141$ & $0.042\pm0.199$ & $0.204\pm0.088$ & $0.295\pm0.121$\\\midrule\multirow{3}{*}{ECG} & LSTM & $0.789\pm0.070$ & $0.873\pm0.056$ & $0.738\pm0.062$ & $0.827\pm0.055$ & $0.954\pm0.007$ & $0.995\pm0.001$\\& TCN & $0.102\pm0.074$ & $0.143\pm0.096$ & $0.072\pm0.062$ & $0.054\pm0.070$ & $0.129\pm0.078$ & $0.189\pm0.101$\\& Transformer & $0.089\pm0.082$ & $0.098\pm0.110$ & $0.020\pm0.092$ & $-0.038\pm0.137$ & $0.315\pm0.066$ & $0.453\pm0.081$\\\midrule\multirow{3}{*}{MS} & LSTM & $0.642\pm0.072$ & $0.768\pm0.066$ & $0.632\pm0.078$ & $0.753\pm0.070$ & $0.950\pm0.007$ & $0.995\pm0.002$\\& TCN & $0.038\pm0.096$ & $0.053\pm0.134$ & $-0.031\pm0.116$ & $-0.058\pm0.167$ & $0.055\pm0.061$ & $0.081\pm0.089$\\& Transformer & $0.124\pm0.136$ & $0.156\pm0.189$ & $0.093\pm0.165$ & $0.076\pm0.244$ & $0.291\pm0.089$ & $0.423\pm0.125$\\
		\bottomrule
	\end{tabularx}   
	\vspace{3mm}
	
	\caption{Consistency Recall@k}\label{tab:recallk}
	\begin{tabularx}{\textwidth}  { XX|ccc|ccc|ccc}
		\toprule
		\multicolumn{2}{c}{}                   & \multicolumn{3}{c}{Feature Permutation (FP)} & \multicolumn{3}{c}{Feature Ablation (FA)} & \multicolumn{3}{c}{Integrated Gradients (IG)} \\ \cmidrule(lr){3-5}\cmidrule(lr){6-8}\cmidrule(lr){9-11}
		\multicolumn{2}{c}{}                   & Top       & Middle       & Bottom       & Top      & Middle      & Bottom      & Top       & Middle       & Bottom       \\ \midrule
		\multirow{3}{*}{PD} & LSTM & $0.041$ & $0.000$ & $0.000$ & $0.072$ & $0.000$ & $0.000$ & $0.165$ & $0.268$ & $0.268$ \\& TCN & $0.227$ & $0.216$ & $0.216$ & $0.454$ & $0.320$ & $0.320$ & $0.103$ & $0.206$ & $0.206$ \\& Transformer & $0.258$ & $0.258$ & $0.258$ & $0.825$ & $0.928$ & $0.928$ & $0.351$ & $0.330$ & $0.330$ \\\midrule\multirow{3}{*}{WIN} & LSTM & $0.064$ & $0.051$ & $0.051$ & $0.103$ & $0.060$ & $0.060$ & $0.244$ & $0.248$ & $0.248$ \\& TCN & $0.256$ & $0.269$ & $0.269$ & $0.500$ & $0.487$ & $0.487$ & $0.286$ & $0.295$ & $0.295$ \\& Transformer & $0.333$ & $0.812$ & $0.812$ & $0.949$ & $0.962$ & $0.962$ & $0.389$ & $0.385$ & $0.385$ \\\midrule\multirow{3}{*}{IPD} & LSTM & $0.250$ & $0.250$ & $0.250$ & $0.625$ & $0.708$ & $0.708$ & $0.375$ & $0.458$ & $0.458$ \\& TCN & $0.250$ & $0.250$ & $0.250$ & $0.875$ & $0.958$ & $0.958$ & $0.292$ & $0.375$ & $0.375$ \\& Transformer & $0.250$ & $0.250$ & $0.250$ & $0.750$ & $0.708$ & $0.708$ & $0.167$ & $0.292$ & $0.292$ \\\midrule\multirow{3}{*}{ECG} & LSTM & $0.171$ & $0.171$ & $0.171$ & $0.341$ & $0.244$ & $0.244$ & $0.256$ & $0.268$ & $0.268$ \\& TCN & $0.256$ & $0.256$ & $0.256$ & $0.634$ & $0.634$ & $0.634$ & $0.329$ & $0.305$ & $0.305$ \\& Transformer & $0.256$ & $0.256$ & $0.256$ & $0.780$ & $0.829$ & $0.829$ & $0.244$ & $0.293$ & $0.293$ \\\midrule\multirow{3}{*}{MS} & LSTM & $0.250$ & $0.262$ & $0.262$ & $0.536$ & $0.560$ & $0.560$ & $0.357$ & $0.321$ & $0.321$ \\& TCN & $0.250$ & $0.262$ & $0.262$ & $0.750$ & $0.798$ & $0.798$ & $0.298$ & $0.381$ & $0.381$ \\& Transformer & $0.250$ & $0.250$ & $0.250$ & $0.845$ & $0.821$ & $0.821$ & $0.274$ & $0.369$ & $0.369$ \\\bottomrule
	\end{tabularx}
\end{table*}

\begin{figure*}[!ht]
	\centering
	\subfloat[Violin plots of Pearson correlation absolute value]{\label{fig:box_rho}\includegraphics[width=\textwidth]{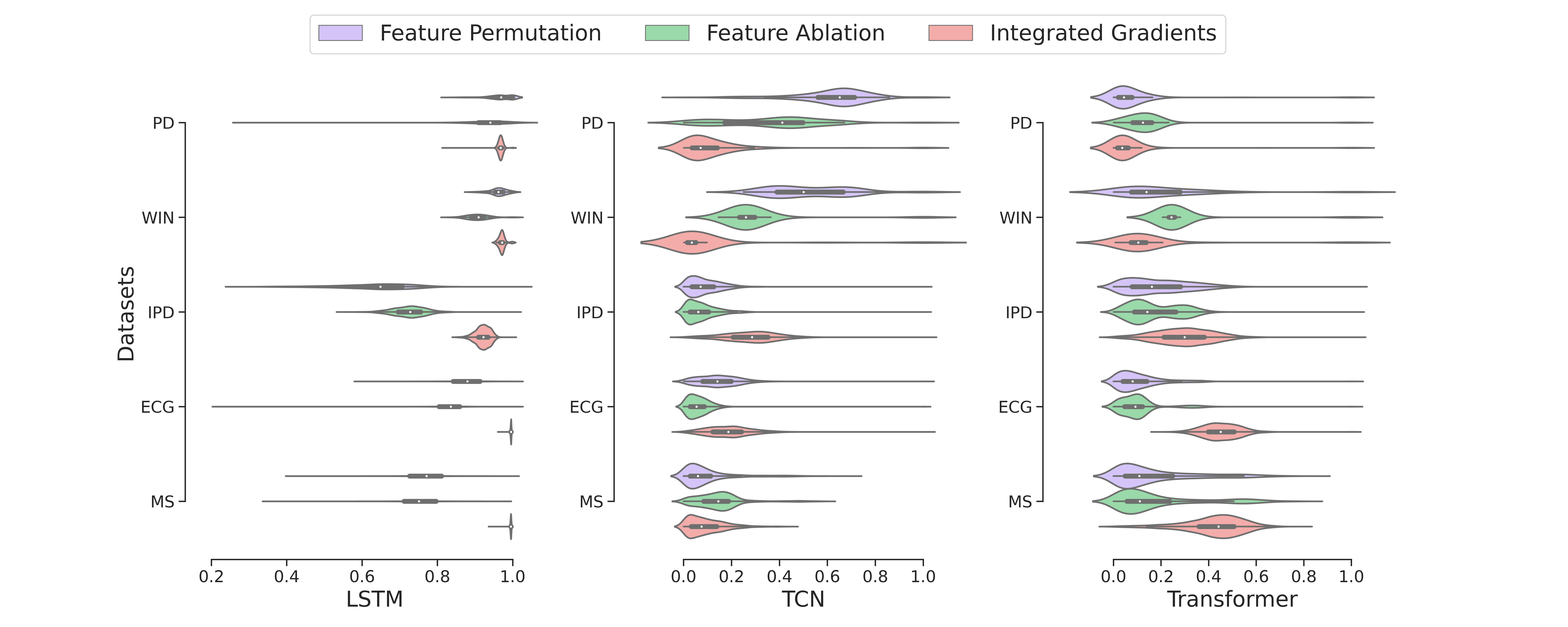} } 
	\hfill
	\subfloat[Violin plots of Kendall's tau absolute value ]{\label{fig:box_tau}    	\includegraphics[width=1.0\textwidth]{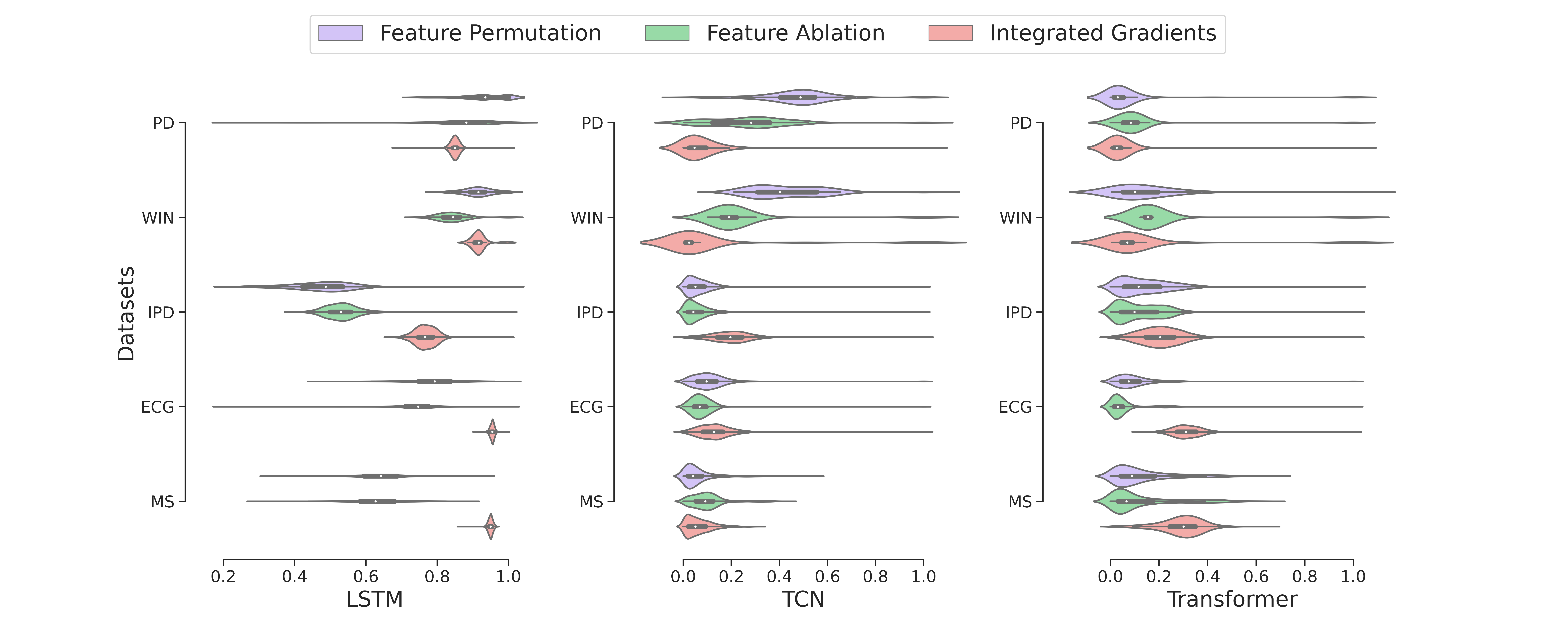}} 
	
	\caption{Each row contains three plots (from left to right, LSTM, TCN, and Transformer architecture), while the colors code for the various explanation methods. Although the two metrics measure something essentially different, the behavior observed in (a) and (b) is similar.}
	\label{fig:tau_rho_consistency}
	
\end{figure*}

\begin{table*}[!ht]
	\centering
	\caption{Robustness ranking analysis}\label{tab:kendalltau_robust}
	
	\begin{tabularx}{\textwidth} {bb|bb|bb|bb}
		\toprule
		\multicolumn{2}{c}{}  
		& \multicolumn{2}{c}{Feature Permutation (FP)} & \multicolumn{2}{c}{Feature Ablation (FA)} & \multicolumn{2}{c}{Integrated Gradients (IG)} \\ 
		\cmidrule(lr){3-4}\cmidrule(lr){5-6}\cmidrule(lr){7-8}
		\multicolumn{2}{c}{}                   & $\tau$     & $\rho$    & $\tau$     & $\rho$    & $\tau$     & $\rho$      \\ \midrule
		\multirow{3}{*}{PD} & LSTM & $0.972\pm0.064$ & $0.985\pm0.036$ & $0.973\pm0.112$ & $0.982\pm0.102$ & $0.807\pm0.064$ & $0.942\pm0.066$\\& TCN & $0.680\pm0.168$ & $0.803\pm0.157$ & $0.597\pm0.259$ & $0.706\pm0.285$ & $0.221\pm0.148$ & $0.307\pm0.175$\\& Transformer & $0.042\pm0.141$ & $0.052\pm0.153$ & $0.129\pm0.148$ & $0.164\pm0.161$ & $0.123\pm0.133$ & $0.171\pm0.143$\\\midrule\multirow{3}{*}{WIN} & LSTM & $0.932\pm0.047$ & $0.971\pm0.021$ & $0.924\pm0.056$ & $0.967\pm0.029$ & $0.929\pm0.075$ & $0.972\pm0.057$\\& TCN & $0.598\pm0.107$ & $0.696\pm0.083$ & $0.553\pm0.094$ & $0.676\pm0.073$ & $0.083\pm0.186$ & $0.105\pm0.189$\\& Transformer & $0.267\pm0.241$ & $0.347\pm0.283$ & $0.466\pm0.153$ & $0.587\pm0.167$ & $0.345\pm0.149$ & $0.473\pm0.143$\\\midrule\multirow{3}{*}{IPD} & LSTM & $0.446\pm0.124$ & $0.603\pm0.147$ & $0.579\pm0.075$ & $0.732\pm0.076$ & $0.699\pm0.062$ & $0.864\pm0.053$\\& TCN & $0.114\pm0.141$ & $0.157\pm0.188$ & $0.277\pm0.159$ & $0.358\pm0.194$ & $0.254\pm0.137$ & $0.361\pm0.183$\\& Transformer & $0.369\pm0.158$ & $0.480\pm0.178$ & $0.364\pm0.215$ & $0.458\pm0.244$ & $0.322\pm0.148$ & $0.449\pm0.192$\\\midrule\multirow{3}{*}{ECG} & LSTM & $0.821\pm0.100$ & $0.898\pm0.073$ & $0.841\pm0.133$ & $0.905\pm0.103$ & $0.967\pm0.029$ & $0.996\pm0.028$\\& TCN & $0.382\pm0.119$ & $0.515\pm0.138$ & $0.517\pm0.087$ & $0.657\pm0.101$ & $0.207\pm0.125$ & $0.300\pm0.172$\\& Transformer & $0.276\pm0.126$ & $0.365\pm0.153$ & $0.535\pm0.141$ & $0.673\pm0.141$ & $0.415\pm0.081$ & $0.582\pm0.101$\\\midrule\multirow{3}{*}{MS} & LSTM & $0.747\pm0.112$ & $0.857\pm0.089$ & $0.694\pm0.107$ & $0.811\pm0.090$ & $0.968\pm0.027$ & $0.996\pm0.024$\\& TCN & $0.147\pm0.140$ & $0.201\pm0.187$ & $0.184\pm0.144$ & $0.248\pm0.189$ & $0.068\pm0.087$ & $0.100\pm0.126$\\& Transformer & $0.473\pm0.220$ & $0.575\pm0.238$ & $0.513\pm0.215$ & $0.627\pm0.239$ & $0.453\pm0.105$ & $0.621\pm0.125$\\
		\bottomrule
	\end{tabularx}    
	\vspace*{3mm}
	\centering
	\caption{Robustness Recall@k}\label{tab:recallk_robust}
	
	\begin{tabularx}{\textwidth} { XX|ccc|ccc|ccc}
		\toprule
		\multicolumn{2}{c}{}                   & \multicolumn{3}{c}{Feature Permutation (FP)} & \multicolumn{3}{c}{Feature Ablation (FA)} & \multicolumn{3}{c}{Integrated Gradients (IG)} \\ \cmidrule(lr){3-5}\cmidrule(lr){6-8}\cmidrule(lr){9-11}
		\multicolumn{2}{c}{}                   & Top       & Middle       & Bottom       & Top      & Middle      & Bottom      & Top       & Middle       & Bottom       \\ \midrule
		\multirow{3}{*}{PD} & LSTM & $0.031$ & $0.000$ & $0.000$ & $0.072$ & $0.000$ & $0.000$ & $0.134$ & $0.237$ & $0.237$ \\& TCN & $0.227$ & $0.258$ & $0.258$ & $0.299$ & $0.443$ & $0.443$ & $0.155$ & $0.216$ & $0.216$ \\& Transformer & $0.268$ & $0.309$ & $0.309$ & $0.742$ & $0.876$ & $0.876$ & $0.268$ & $0.299$ & $0.299$ \\\midrule\multirow{3}{*}{WIN} & LSTM & $0.056$ & $0.030$ & $0.030$ & $0.103$ & $0.051$ & $0.051$ & $0.231$ & $0.248$ & $0.248$ \\& TCN & $0.274$ & $0.265$ & $0.265$ & $0.389$ & $0.355$ & $0.355$ & $0.295$ & $0.282$ & $0.282$ \\& Transformer & $0.303$ & $0.252$ & $0.252$ & $0.466$ & $0.568$ & $0.568$ & $0.179$ & $0.372$ & $0.372$ \\\midrule\multirow{3}{*}{IPD} & LSTM & $0.250$ & $0.250$ & $0.250$ & $0.458$ & $0.417$ & $0.417$ & $0.375$ & $0.458$ & $0.458$ \\& TCN & $0.250$ & $0.250$ & $0.250$ & $0.458$ & $0.500$ & $0.500$ & $0.333$ & $0.375$ & $0.375$ \\& Transformer & $0.250$ & $0.250$ & $0.250$ & $0.458$ & $0.542$ & $0.542$ & $0.292$ & $0.333$ & $0.333$ \\\midrule\multirow{3}{*}{ECG} & LSTM & $0.244$ & $0.134$ & $0.134$ & $0.341$ & $0.232$ & $0.232$ & $0.256$ & $0.280$ & $0.280$ \\& TCN & $0.256$ & $0.256$ & $0.256$ & $0.476$ & $0.476$ & $0.476$ & $0.232$ & $0.256$ & $0.256$ \\& Transformer & $0.256$ & $0.256$ & $0.256$ & $0.634$ & $0.622$ & $0.622$ & $0.390$ & $0.305$ & $0.305$ \\\midrule\multirow{3}{*}{MS} & LSTM & $0.250$ & $0.250$ & $0.250$ & $0.381$ & $0.417$ & $0.417$ & $0.333$ & $0.262$ & $0.262$ \\& TCN & $0.250$ & $0.250$ & $0.250$ & $0.500$ & $0.500$ & $0.500$ & $0.405$ & $0.381$ & $0.381$ \\& Transformer & $0.250$ & $0.250$ & $0.250$ & $0.357$ & $0.405$ & $0.405$ & $0.190$ & $0.226$ & $0.226$ \\\bottomrule
	\end{tabularx}
\end{table*}

\subsection{Datasets}
We consider five real-world univariate time series datasets: \emph{Power Demand} (PD), \emph{Wine} (WIN), \emph{Italy Power Demand} (IPD), \emph{Two Lead ECG} (ECG) and \emph{Mote Strain} (MS). PD derives from Keogh et al.~\cite{keogh2005hot}, while the others are available in the UCR Archive~\cite{UCRArchive}. We preprocess all datasets by dividing them into non-overlapping windows a priori, and the class labels of each window are available. However, the ground truth data does not include the attribution of the prediction. In~\Cref{sec:exp_consistency,sec:exp_robust}, we introduce artificial padding with random noise to each input window and assign equal importance to the area of the original input.

\subsection{Experimental setup}
For our experiments, we select three representatives from the common saliency explanation approaches~\cite{ismail2020benchmarking} for time series data. We employ \emph{Feature Permutation} (FP) and \emph{Feature Ablation}(FA)~\cite{suresh2017clinical}, which are perturbation-based methods,  and \emph{Integrated Gardients} (IG)~\cite{sundararajan2017axiomatic}, which is a gradient-based method. We use the implementation provided by Ismail et al.~\cite{ismail2020benchmarking}. \par
We investigate the behavior of saliency explanations on three types of network structures Recurrent Neural Networks (RNNs), Convolutional Neural Networks (CNNs), and attention-based networks. To this end, we picked three implementations commonly used for time series data: LSTM~\cite{hochreiter1997long}, TCN~\cite{lea2017temporal}, and Transformer~\cite{vaswani2017attention}. We configure these models with a Softmax output layer for classification and train the models on all the padded variants of the input windows, including top, middle and bottom padding. During the test phase, we generate saliency maps and analyze the effect of each group of padding variants separately. 

\subsection{Consistency evaluation}\label{sec:exp_consistency}
We apply artificial padding to each univariate time window to evaluate the explanation consistency over sliding windows. Specifically, we expand each univariate data window $w_s^d\in\mathbb{R}^{n\times d}$ to a matrix $m\in \mathbb{R}^{\alpha\times \lfloor\beta \cdot d\rfloor} (\alpha>m, 1<\beta<3)$. The data window $w_s^d$ is placed on $d$ consecutive dimensions of $m$, and the other dimensions are filled with randomly sampled noise from a normal distribution. The effect of a sliding window can be simulated by placing $w_s^d$ at different rows in $m$. Specifically, we allocate $w_s^d$ at the top, middle, and bottom third of $m$ to generate three overlapping sliding windows, i.e., three variants of each input window. We call the area in the saliency map corresponding to the input window $w_s^d$ , the \emph{area of interest}. We show the experimental results by setting $\alpha=4$ and $\beta=\frac{5}{3}$. An example of the padded data window is shown in \Cref{fig:saliencymaps}. \par
The proposed construction we get that each padding variant group (top/middle/bottom) contains the same input window only located differently. To examine the consistency of the saliency explanations, we compare the feature ranking of the obtained attributions in corresponding locations in each padding variant. As a showcase, we visualize the result of one window from the IPD dataset in~\Cref{fig:saliencymaps}. The left three columns of Figure~\ref{fig:saliencymaps} represent the saliency explanations in the various padded input windows, the $y$-axis being the time and the $x$-axis being the input features. Only the second feature contains essential information to be learned by the classifiers (see the first column in \Cref{fig:saliencymaps}). The saliency maps on the right side correspond to the three explanation models FP, FA, and IG. We expect the second feature column's top, middle, and bottom third to be marked as salient. However, as Ismail et al.~\cite{ismail2020benchmarking} have already shown, classical saliency methods might fail on time series data due to the temporal feature, and our experiments confirm their results (see~\Cref{fig:saliencymaps}) where the latest timestamps play more important roles in the prediction. The various explainers can detect the important timestamps and suffer from distinguishing important features for TCN and Transformers. \par 
Despite the sub-optimal saliency explanations, we analyze the consistency between the padding variants. We evaluated the disagreement empirically on the saliency explanations using Kendall's $\tau$~\cite{kendall1948rank} and Pearson correlation; Kendall's $\tau$  measures the smallest number of swaps of adjacent elements that transform one ranking into the other while the Pearson correlation coefficient measures the covariance of the two random variables divided by the product of their standard deviations. All quantities can be estimated using finite samples.\par
We calculate the importance scores for each timestamp and input feature, obtaining the importance ordering of the \emph{area of interest}. For each pair of ranking from the three padding variants, we analyzed the pairwise comparisons among rankings of feature-time pixels in the saliency explanations FP, FA, and IG. The average Kendall's $\tau$ and Pearson correlation ($\rho$) are summarized in~\Cref{tab:kendalltau_std} and the absolute values are visualized in~\Cref{fig:tau_rho_consistency}.\par
Table~\ref{tab:kendalltau_std} contains, for each data set, neural network architecture, and saliency map, the average Kendall's $\tau$ and Pearson correlation coefficients with the respective variance. From the table, it is easy to spot how the importance scores rankings provided vary in ranges below $1$. Kendall's $\tau$ and Pearson correlation coefficients range between $1$ and $-1$, where $1$ indicates complete agreement among the rankings, while values close to zero suggest non-constant and independent orderings. From~\Cref{fig:box_rho,fig:box_tau}, we observe that the coefficients are, in most cases, crowded at low values, and episodes of perfect agreement among the obtained rankings in the different windows are rare (although non-anomalies). \par
A special case is the LSTM algorithm that provides consistent saliency maps among the various windows. However, observing the explanation over the LSTM model, we see that both Kendall's $\tau$ and Pearson correlation coefficients tend to accumulate to high scores ($\approx 1$) as the LSTM method tends to accumulate the learning in the last timestamps, thus implying that the explanation methods assign high importance only to the last timestamps. We further observe FA correctly finds the relevant timestamps but cannot distinguish between noisy and relevant features. \par
In addition to the relative ranking, we also check the quality of the saliency explanation using Recall$@k$. Table~\ref{tab:recallk} contains the Recall$@k$ obtained among the importance rankings of timestamps in the \emph{areas of interest}. Recall$@k$ measures the ratio among correctly relevant and retrieved elements and the number of relevant elements and ranges in $[0,1]$. High recall ($\approx 1$) indicates that the highly ranked feature-time pixels are concentrated in the area of interest, while low Recall$@k$ indicates the inability to find relevant elements correctly.

\subsection{Robustness evaluation}
\label{sec:exp_robust}
To evaluate the robustness of the saliency explanation, we apply the feature swapping depicted in~\Cref{fig:swaps}. Specifically, we continue using the padded input matrix $m\in \mathbb{R}^{\alpha\times \lfloor\beta \cdot d\rfloor}$ from \Cref{sec:exp_consistency} and swap the feature dimensions containing the original input data window (\emph{area of interest}) with noise dimensions. We train different classification models with the swapped and not swapped data. For simplicity, we always locate the original window in the middle of the selected feature dimension in this experiment. We compare the ranking of feature-time pixel explanations in the \emph{areas of interest} of swapped and not swapped pairs. The Kendall's $\tau$ and Pearson correlation $\rho$ are summarized in~\Cref{tab:kendalltau_robust}.
\par
The absolute values of Kendall's $\tau$ and Pearson correlation for TCN and Transformers indicate a significant difference in saliency maps after the swapping. In other words, when the important feature is switched with a noisy feature, the feature attribution in the saliency map is not switched correspondingly. An exception is the LSTM classifier, which explains all datasets except IPD robustly. However, the explanation quality is limited.

\section{Conclusions}
While explanations based on saliency maps have succeeded in vision and natural language domains, they remain still challenging for time series data. In addition to the well-known challenges posed by the additional time dimension to the input features, we have also identified issues related to \emph{consistency} and \emph{robustness}.\par
We exploited both the issue of \emph{inconsistency} raising in saliency explanation over overlapping time windows and the issue of \emph{non-robustness} when swapping features in time series windows. The presented exploratory analysis aims to raise awareness of the described problems and motivates further development of saliency methods that address the existing flaws. 

\section*{Acknowledgments}
{This research was supported by the research training group \emph{Dataninja} funded by the German federal state of North Rhine-Westphalia and by the Research Center Trustworthy Data Science and Security, an institution of the University Alliance Ruhr.}

\bibliographystyle{ACM-Reference-Format}
\bibliography{main}

\end{document}